\documentclass[wcp]{jmlr}

\usepackage{longtable}

\RequirePackage{algorithm}
\RequirePackage{algorithmic} 
\usepackage{color, colortbl}

\usepackage{booktabs}

\makeatletter
\let\Ginclude@graphics\@org@Ginclude@graphics 
\makeatother

\jmlrvolume{}
\jmlryear{2022}
\jmlrworkshop{ACML 2022}
\jmlrvolume{189}
\editors{Emtiyaz Khan and Mehmet G\"{o}nen}
\title[ITM-VAE]{Interpretable Representation Learning from Temporal Multi-view Data}

  \author{\Name{Lin Qiu} \Email{lin.qiu.stats@gmail.com}\\
   \addr The Pennsylvania State University
   \AND
   \Name{Vernon M. Chinchilli} \Email{vchinchi@psu.edu} \\
    \addr The Pennsylvania State University
   \AND
   \Name{Lin Lin} \Email{l.lin@duke.edu}\\
  \addr Duke University}

\begin{document}

\maketitle

\begin{abstract}
In many scientific problems such as video surveillance, modern genomics, and finance, data are often collected from diverse measurements across time that exhibit time-dependent heterogeneous properties. Thus, it is important to not only integrate data from multiple sources (called multi-view data), but also to incorporate time dependency for deep understanding of the underlying system. We propose a generative model based on variational autoencoder and a recurrent neural network to infer the latent dynamics for multi-view temporal data.  This approach allows us to identify the disentangled latent embeddings across views while accounting for the time factor. We invoke our proposed model for analyzing three datasets on which we demonstrate the effectiveness and the interpretability of the model.
\end{abstract}
\begin{keywords}
temporal multi-view data; model interpretability; variational inference
\end{keywords}

\section{Introduction}
Multi-view data is prevalent in real-world applications. For instance, a photo can be taken at different angles, the human motion can be described by different gestures, and medical scenarios where each observed clinical outcome of a patient can correspond to a specific medical test. These views often represent diverse and complementary information of the same data. Integrating multi-view data has the potential to yield more generalizable representations and is helpful in boosting the performance of data mining tasks. 

Temporal multi-view data arise in a wide variety of fields, such as biomedical research, sociology, finance, computer vision and many others \citep{yang17} in which datasets are collected repeatedly over time for each individual. Analyzing temporal multi-view data in such studies, with the objective of delivering interpretable learning, is challenging.

Many popular multi-view learning methods have been developed based on group factor analysis~\citep{Sridharan08,Klami15,zhao16,Leppaaho17}, where each group corresponds to a specific data view. The group factor analysis generates a common linear mapping between the latent and observed groups of variables (multiple views). In order to further extract interpretable information, most of the methods exploit the idea of using sparse linear factor models. In particular, the resulting latent factor is restricted to contribute to variation in only a subset of the observed features. For example, sparse factor loadings in gene expression data analysis can be interpreted as non-disjoint clusters of co-regulated genes~\citep{Pournara07,Lucas10,Gao13}.  
Limitations of those existing sparse methods for the application in real data scenarios are scalability and the inability to handle nonlinear time-dependent complex structures~\citep{Ainsworth18}. 

The variational auotencoders (VAE) is a powerful deep generative learning technique for efficient inference~\citep{Kingma14,rezende14}. VAE learns a lower dimensional representation of the data through an encoder, and then with a decoder transforms the latent representation back to the original data space. VAE is flexible and can account for any kind of data. However, the original VAE ignores the temporal correlations across latent dimensions.

\textbf{Contributions} Our motivation lies in the study of high-dimensional temporal multi-view data. We seek to infer trajectories of latent variables that provide insights into the latent, lower-dimensional structure derived from the dynamics of the observed data space. Motivated by the success of variational recurrent neural network (VRNN) for modeling temporal sequence data, we propose a new modeling strategy that integrates VRNN into sparse group factor analysis. We label this model as the interpretable representation learning for temporal multi-view data (ITM-VAE).
The resulting model, thus serves as a nonlinear factor model for multi-view data observed across time. Our main contributions can be summarized as follows: 

\begin{itemize}
    \item  We build a novel interpretable model that can perform sensible disentanglement for temporal multi-view data. The ability of ITM-VAE to perform sensible disentanglement is through the introduction of a view and time-specific transformation  $\mathbf{W}$ (introduced in Section~\ref{sec:ITM-VAE}) which is built on a sparsity inducing prior between the time-specific latent representation and the view and time-specific neural generator. 
    \item We derive an efficient timestep-wise variational inference scheme for learning temporal multi-view data.

    \item We show that ITM-VAE can learn dynamic dependency among views. This is an appealing feature of ITM-VAE because most of the complex systems depend on a temporal component, and such a component contributes to the development of variable interactions gradually. The ability to access the dynamic relationship of groups of variables will help us gain insight for downstream analysis of the complex data. 
\end{itemize}

\section{Related Work}

A few extensions have been proposed for VAE to model the correlations in the latent space. The conditional VAE (CVAE)~\citep{Sohn15} is a graphical model,  and its input observations modulate the prior on Gaussian latent variables that generate the outputs. However, CVAE cannot model the individual sample-specific temporal structure~\citep{Sohn15}. GPPVAE~\citep{casale18} combines the VAE and the Gaussian process (GP) prior over the latent space to model the temporal dependencies between samples. Due to the restrictive nature of the view-object GP product kernel, GPPVAE cannot capture the individual-specific temporal structure. An extension of GPPVAE, GP-VAE \citep{fortuin20}, is designed specifically for data imputation. GP-VAE places an independent GP prior on each individual sample's time-series to relax the inference technique. A limitation of GP-VAE is that GP-VAE cannot capture the shared temporal structure across all data points. DP-GP-LVM is a nonparametric Bayesian latent variable model that aims to learn the dependency structures of multimodal data by the GP prior~\citep{Lawrence19}. GP prior is shown to be well suited for time series modeling, however, it comes at the cost of inverting the kernel matrix, which has a time complexity of $\mathcal{O}(d^3)$, where $d$ is the dimensionality of the data. Moreover, it  is often a challenge to design a kernel function that can accurately capture both the correlation in feature space and in a temporal dimension. 

The recurrent neural networks (RNNs)~\citep{Martens11,Hermans13,Pascanu13,Graves13} have shown good performance in modeling sequence data, where the latent random variables in the RNN function serve as ``memory" of the past sequence.  
RNN can be further extended to integrate the dependencies between the latent random variables at neighboring time steps, called variational recurrent neural network (VRNN)~\citep{Chung15} in the context of VAE. VRNN can handle complex nonlinear and highly structured sequential data, 

Output interpretable VAE (oi-VAE) is designed for non-temporal grouped data with a structured VAE comprised of group-specific generators \citep{Ainsworth18}. The latent variables are shared across all groups and are assumed to be $iid$ for each data point. Because of the group design of oi-VAE, its model interpretation is limited to factor level. Our proposed ITM-VAE is a temporal extension of oi-VAE with a feature level interpretation.


\section{Background}

\begin{figure}[t!]
\vspace{-0.05in}
\begin{center}
\centerline{\includegraphics[width=1.2in]{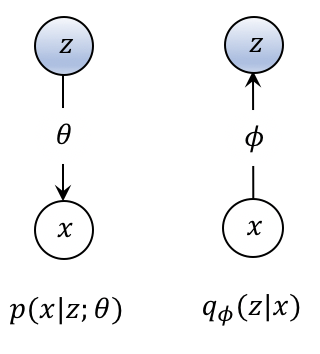}}
\vspace{-0.05in}
\caption{Generative Model. Left: The conditional probability $p(x|z;\theta)$ parameterized by a non-linear deep neural network through the latent variable $z$. Right: The inference network $q_\phi(z|x)$.}
\label{fig:generative_model}
\end{center}
\vspace{-0.3in}
\end{figure}

\textbf{Generative Model}:
In generative models as shown in Fig.~\ref{fig:generative_model}, the class of VAEs are popular for efficient approximate inference and learning~\citep{Kingma14}. 
VAE approximates intractable posterior distributions over latent representations that are parameterized by a deep neural network, which maps observations to a distribution over latent variables.

For non-sequential data, VAE has become one of the most popular approaches for efficiently recovering complex multimodal distributions. Recently, VAE has been extended to dynamic systems~\citep{Archer15}. Briefly, VAE provides a mapping from the observations to a distribution on their latent representation. The resulting simpler latent subspace can be used to describe the underlying complex system. Mathematically, let $\mathbf{x}\in \mathcal{R}^d$ denote a $d$-dimensional observation and $\mathbf{z}\in \mathcal{R}^k$ denote a vector of latent random variables of fixed dimension $k$ with $k < d$. The generative process of VAE can be represented as:
\begin{align}
\mathbf{z}\sim \mathcal{N}(0, I), \mathbf{x}\sim \mathcal{N}(\boldsymbol{\mu}_\mathbf{x}, D), 
\end{align}
where $I$ is the identity matrix, $D$ is a $d\times d$ diagonal matrix whose diagonals are the marginal variances of each component of $\mathbf{x}$, and $\boldsymbol{\mu}_\mathbf{x}$ is the mean of the Gaussian likelihood which is produced by a neural network with parameters $\theta$ taking $\mathbf{z}$ as an input. 
Then the joint distribution is defined as:
\begin{align}
    \label{eq:bayes}
    p(\mathbf{x}, \mathbf{z};\theta) = p(\mathbf{x} \mid \mathbf{z};\theta)p(\mathbf{z}).
\end{align}

\textbf{Learning Inference}:
One unique feature of VAE is that it allows the conditional distribution $p(\mathbf{x} \mid \mathbf{z})$ to be a potentially highly nonlinear mapping from $\mathbf{z}$ to $\mathbf{x}$.

The likelihood is then parameterized with a generative network (called decoder). VAE uses $q(\mathbf{z}| \mathbf{x})$ with an inference network (called encoder) to approximate the posterior distribution of $\mathbf{z}$. For example, $q(\mathbf{z}|\mathbf{x})$ can be a Gaussian $\mathcal{N}(\mu, \sigma^2I)$, where both $\mu$ and $\sigma^2$ are parameterized by a neural network: $[\mu, \log \sigma^2] = f_\phi(\mathbf{x})$, where $f_\phi$ is a neural network with parameters $\phi$.
The parameters for both generative and inference networks are learned through variational inference, and Jensen's inequality yields the evidence lower bound (ELBO) on the marginal likelihood of the data:

\begin{equation}
\begin{split}
\label{eqn:varlowbnd}
\log p_\theta(\mathbf{x})\ge
\underbrace{\mathbb{E}_{q(\mathbf{z}; \phi)}[\log p_\theta(\mathbf{x}\mid \mathbf{z})] - \mathrm{KL}(q(\mathbf{z}; \phi)\mid\mid p(\mathbf{z})),}_{
\mathcal{L}(x; \theta, \phi)}
\end{split}
\end{equation}    
  where $\mathrm{KL}(Q \| P)$ is the Kullback-Leibler divergence between two distributions $Q$ and $P$. $q(\mathbf{z};\phi)$ is a tractable variational distribution meant to
approximate the intractable posterior distribution $p(\mathbf{z}\mid \mathbf{x})$; it is controlled by some parameters $\phi$.
We want to choose $\phi$ that makes the bound in Eq.~\eqref{eqn:varlowbnd} as tight
as possible, $\phi^* \triangleq \arg\max_\phi \mathcal{L}(\mathbf{x}; \theta, \phi)$.

One can train a feedforward
inference network to find good variational parameters $\phi(\mathbf{x})$ for a given $\mathbf{x}$, where $\phi(\mathbf{x})$ is the output of a neural network with parameters $\phi$ that are trained
to maximize $\mathcal{L}(\mathbf{x}; \theta, \phi(\mathbf{x}))$~\citep{Kingma14}.

\section{ITM-VAE model}\label{sec:ITM-VAE}

Let $\mathbf{x}_{1:T} = (\mathbf{x}_1,...,\mathbf{x}_T)$ denote the sequence data observed at $T$ timesteps. We then rewrite $\mathbf{x}_t$, which is the sequence data at timestep $t$,  to incorporate the view information as $\mathbf{x}_t = [\mathbf{x}^{(1)}_t,..., $ $\mathbf{x}^{(G)}_t] \in \mathcal{R}^{d_1 + d_2,...+ d_G}$, where $\mathbf{x}^{(g)}_t$ denote the data from the $g$th view at time $t$ and $G$ is the number of views, each view has the dimension of $d_g$.

\subsection{Modeling Framework}

\textbf{Prior}
Given a temporal sequence of vectors $\mathbf{x}_{1:T} = (\mathbf{x}_1,...,\mathbf{x}_T)$, $\mathbf{x}_t \in \mathcal{R}^d$,
the conventional VAEs assume an independent latent variable $\mathbf{z}$ for each timestep $t$: $\mathbf{z}\sim  \mathcal{N}(0, I)$. To encode temporal variability, we propose to allow the latent variable $\mathbf{z}_t$ at timestep $t$ to depend on the  state variable $\mathbf{h}_{t-1}$ of an RNN though the following distribution: 

\begin{align}
    \label{eq:vrnn_prior}
    \mathbf{z}_t \sim \mathcal{N}(\boldsymbol{\mu}_{0,t}, \text{diag}(\boldsymbol{\sigma}_{0,t}^2)), \;\;
    [\boldsymbol{\mu}_{0,t}, \boldsymbol{\sigma}_{0,t}] = \varphi_{\tau}^{\text{prior}} (\mathbf{h}_{t-1}),
\end{align}
where both $\boldsymbol{\mu}_{0,t}$ and $\boldsymbol{\sigma}_{0,t}$ are produced by a distinct neural network that approximates the time-dependent prior distribution~\citep{Chung15}. More specifically,  $[\boldsymbol{\mu}_{0,t}, \boldsymbol{\sigma}^2_{0,t}] = \varphi_{\tau}^{\text{prior}} (\mathbf{h}_{t-1})$, and $\varphi_{\tau}^{\text{prior}}(\mathbf{h}_{t-1})$ denotes a neural network taking the previous hidden state $\mathbf{h}_{t-1}$ as input.

\textbf{Encoder}
Similar to the VAEs, we need to define an approximate posterior $q(\mathbf{z}|\mathbf{x})$. We propose to let $\mathbf{z}_t$ capture the shared variability among views at each timestep by allowing $q(\mathbf{z}|\mathbf{x})$ as a function of both $\mathbf{x}_t = [\mathbf{x}^{(1)}_t,..., $ $\mathbf{x}^{(G)}_t]$ and $\mathbf{h}_{t-1}$ as:
\begin{align}
\label{vnn:enc}
\mathbf{z}_t |\mathbf{x}_t \sim \mathcal{N}(\boldsymbol{\mu}_{\mathbf{z},t}, \boldsymbol{D}_{\mathbf{z},t}), \;\;  \\ 
[\boldsymbol{\mu}_{\mathbf{z},t}, 
\text{diag}(\boldsymbol{D}_{\mathbf{z},t})] = \varphi_{\tau}^{\text{enc}}(\varphi_{\tau}^{\mathbf{x}}(\mathbf{x}_t), \mathbf{h}_{t-1}).
\end{align}

\textbf{Decoder} The generation of $\mathbf{x}_t$
will depend on both $\mathbf{z}_t$ and $\mathbf{h}_{t-1}$. In addition, we propose to model different views of data independently while allowing the latent variable $\mathbf{z}_t$ to be shared across $G$ views at timestep $t$. The corresponding generative distribution will be:
\begin{align}
    \label{eq:vrnn_gen}
    \mathbf{x}^{(g)}_t \mid \mathbf{z}_t \sim
    \mathcal{N}(\boldsymbol{\mu}^{(g)}_{\mathbf{x},t} ,
   \boldsymbol{D}^{(g)}_{\mathbf{x},t}),
\end{align}
where $\mathbf{D}^{(g)}_{\mathbf{x},t}$ is a diagonal matrix.   We then introduce a sequence of latent matrices $\mathbf{W}^{(g)}_t \in \mathcal{R}^{d_g\times k}$, for $t=1:T, g = 1:G$. $\mathbf{W}$ will help with model interpretation by placing a column-wise sparsity prior which will be introduced in section \ref{method_interpret}: Model Interpretability. Both parameters $\boldsymbol{\mu}^{(g)}_{\mathbf{x},t}$ and $\mathbf{D}^{(g)}_{\mathbf{x},t}$ will be conditioned on $\mathbf{W}^{(g)}_t$, $\mathbf{z}_t$ and $\mathbf{h}_{t-1}$ through: 
\begin{align}
    \label{eq:vrnn_gen1}
[\boldsymbol{\mu}^{(g)}_{\mathbf{x},t}, \text{diag}(\boldsymbol{D}^{(g)}_{\mathbf{x},t}))] = \varphi_{\theta_{t,g}}^{\text{dec}}(\mathbf{W}^{(g)}_t \mathbf{z}_t, \mathbf{h}_{t-1}),\end{align}
where $\varphi_{\theta_{t,g}}^{\text{dec}}$ denotes a neural network with parameters $\theta_{t,g}$, and $\text{diag}(\mathbf{D})$ denotes the diagonal elements of the matrix $\mathbf{D}$. 

\textbf{Recurrence}
The hidden state $\mathbf{h}_t$ is updated by conditioning on $\mathbf{z}_t$ in a recurrent way: $\mathbf{h}_t = S_{\theta}\left(\mathbf{x}_t, {\mathbf{z}_t}, \mathbf{h}_{t-1}\right)$, where $S$ is the transition function which can be implemented with gated activation functions such as long short-term memory or gated recurrent unit \citep{cho14,hochreiter97}. VRNN demonstrates that  including feature extractors in the recurrent equation is important for learning complex data:
\begin{align}
\label{eq:vrnn_rec}
\mathbf{h}_t = S_{\theta}\left(\varphi_{\tau}^{\mathbf{x}}(\mathbf{x}_t), \varphi_{\tau}^{\mathbf{z}}(\mathbf{z}_{t}), \mathbf{h}_{t-1}\right),
\end{align}
where $\varphi_{\tau}^{\mathbf{x}}$ and $\varphi_{\tau}^{\mathbf{z}}$ are two neural networks for feature extraction from $\mathbf{x}_t$ and $\mathbf{z}_t$, respectively. 
By the above model specifications, the generative distribution can be factorized as: 
\begin{align}
p(\mathbf{x}_{1:T}, \mathbf{z}_{1: T}) = \prod_{t=1}^{T} [\prod_{g = 1}^G  p(\mathbf{x}^{(g)}_t | \mathbf{z}_{\leq t}, \mathbf{x}^{(g)}_{\leq t})]p(\mathbf{z}_t| \mathbf{x}_{\leq t}, \mathbf{z}_{\leq t}).
\end{align}

\begin{figure*}[ht]
\vskip -0.05in
\centering
\vspace{.3in}
\includegraphics[width=4.5in]{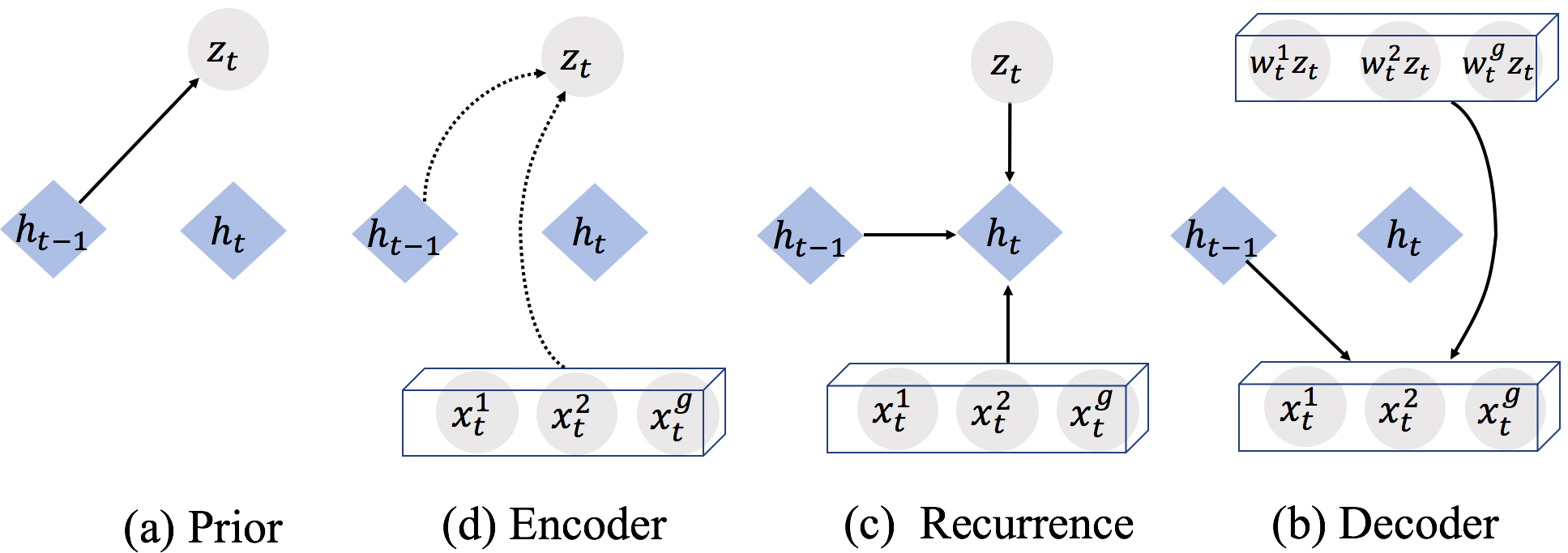}
\caption{Graphical illustration of each operation in ITM-VAE model. Light blue color represents the hidden states. Grey color represents data (in the box) and latent variable. {(a)} Computing the conditional prior using Eq.~\eqref{eq:vrnn_prior}; {(b)} Inference of the approximate posterior using Eq.~\eqref{vnn:enc}; (c) Updating RNN hidden states using Eq.~\eqref{eq:vrnn_rec}, and {(d)} Generating function using Eqs.~\eqref{eq:vrnn_gen} and~\eqref{eq:vrnn_gen1}.}
\label{fig:concept}

\end{figure*}

The ITM-VAE model structure is depicted in Fig.~\ref{fig:concept}.

\textbf{Model Interpretability} \label{method_interpret} Similar to oi-VAE \citep{Ainsworth18}, we  use a spare prior \citep{Kyung10} on the weight matrix to achieve interpretable results.
We place a column-wise sparsity prior for $\mathbf{W}^{(g)}_t$ to ensure the interpretability of the model: 
\begin{align}
\gamma_{gjt}^2 \sim \text{Gamma}\left(\frac{d_g + 1}{2}, {\lambda^2}/{2}\right), \\
\mathbf{W}^{(g)}_{t\boldsymbol{:}, j} \sim \mathcal{N}(0, \gamma_{gjt}^2 I),
\end{align}
where $d_g$ denotes the number of features for each view $g$. Note that our prior specification  is different from oi-VAE,  which assumes each view's input dimensions are the same, and that is the reason why oi-VAE is limited in factor level interpretation and cannot track the feature sparsity for downstream analysis. 
The parameter $\lambda$ controls the model sparsity. More specifically, a larger value of $\lambda$ implies more column-wise sparsity in $\mathbf{W}_{t\boldsymbol{:},j}^{(g)}$. Marginalizing over $\gamma_{gjt}^2$ induces group sparsity over the columns of $\mathbf{W}_{t\boldsymbol{:},j}^{(g)}$. 
Hence, the model automatically tracks the sparse features among groups through time.

\subsection{Timestep-wise Learning}

The traditional VAEs are learned by optimizing  the ELBO using stochastic gradient methods. We are more interested in the sparsity of the learned $\mathbf{W}_{t\boldsymbol{:},j}^{(g)}$ for model interpretability.  
The sparsity inducing prior on $\mathbf{W}_{t\boldsymbol{:},j}^{(g)}$ is marginally equivalent to the convex group lasso penalty. Hence, we propose to adapt the idea of collapsed variational inference~\citep{Ainsworth18} to obtain the true sparsity of the columns $\mathbf{W}_{t\boldsymbol{:},j}^{(g)}$, and apply the timestep-wise variational lower bound.

Let $\mathcal{W} = \left({\mathbf W}^{(1)}_{1:T}, \cdots, {\mathbf W}^{(G)}_{1:T}\right)$, $\gamma^2 =\left(\gamma^2_{1:G, 1:K, 1:T}\right)$, $\mathbf{x} = \mathbf{x}_{1:T}$, and $\mathbf{z} = \mathbf{z}_{1:T}$.

We can compute $\log p(\mathbf{x})$ by marginalizing out all $\gamma_{gjt}^2$'s:
\begin{equation*}
\begin{split}
&\log p(\mathbf{x}) = \log \int \int p(\mathbf{x} | \mathbf{z}, \mathcal{W}, \theta) p(\mathbf{z}) p(\mathcal{W} | \gamma^2) p(\gamma^2) p(\theta) \,d\gamma^2 \,dz \\
&= \log \int \left( \int p(\mathcal{W}, \gamma^2) \,d\gamma^2 \right) \frac{p(\mathbf{x} | \mathbf{z}, \mathcal{W}, \theta) p(\mathbf{z}) p(\theta)}{q_\phi(\mathbf{z} | \mathbf{x}) / q_\phi(\mathbf{z} | \mathbf{x})} \,dz \\
&\geq \mathbb{E}_{q_\phi(\mathbf{z}_{\leq T}\mid\mathbf{x}_{\leq T})}\Bigg[\sum_{t=1}^T -\mathrm{KL}(q_\phi(\mathbf{z}_{t} \mid \mathbf{x}_{\leq t}, \mathbf{z}_{< t}) \| \\ & p(\mathbf{z}_t \mid \mathbf{x}_{< t}, \mathbf{z}_{< t})) + \log p(\mathbf{x}_t \mid \mathbf{z}_{\leq t}, \mathbf{x}_{< t}) \Bigg] \\
&+ \log p(\theta_t) - \lambda \sum_{t=1}^T \sum_{g,j} || \mathbf{W}_{t\boldsymbol{:},j}^{(g)} ||_2,
\end{split}
\end{equation*}
where $\gamma_{gjt}^2 \sim \text{Gamma}\left(\frac{d_g + 1}{2}, {\lambda^2}/{2}\right)$, 
$\mathbf{W}^{(g)}_{t\boldsymbol{:}, j} \sim \mathcal{N}(0, \gamma_{gjt}^2 I)$, and $\phi$, $\theta$ are neural network parameters.

\subsection{Optimization}

\citet{parikh14} proposed the proximal gradient descent algorithms which are a broad class of optimization techniques for separable objectives with both differentiable and potentially non-differentiable components,
\begin{align}
\min_{x} g(x) + h(x),
\end{align}
where $g(x)$ is differentiable and $h(x)$ is potentially non-smooth or non-differentiable. oi-VAE \citep{Ainsworth18} stated that collapsed variational inference with proximal updates provided faster convergence and succeeded in identifying sparser models than other techniques. We chose to use proximal gradient descent updates on our temporal latent-to-group matrices $\text {\bf W}_{t{:},j}^{(g)}$ for timestep-wise learning like oi-VAE did on their latent-to-group matrices $\text {\bf W}_{.j}^{(g)}$.


In our scenario, we define
\begin{align}
\mathbf{x}^{s+1} = \text{prox}_{\lambda^s g} (\mathbf{x}^s - \lambda^s \nabla f(\mathbf{x}^s)),
\end{align}
where $s$ denotes the $s$th iteration, $\lambda^s > 0$ is a step size, and $\text{prox}_f (\mathbf{x})$ is the proximal operator for the function $f$. Expanding the definition of $ \text{prox}_{\lambda^s g}$, we can show that the proximal step corresponds to minimizing $g(\mathbf{x})$ plus a quadratic approximation to $g(\mathbf{x})$ centered on $\mathbf{x}^s$.
For $f:\textbf{R}^n \rightarrow \textbf{R}$ and $g:\textbf{R}^n \rightarrow \textbf{R} \cup \{+\infty\}$ are closed proper convex and $f$ is differentiable.
~For $g(\text{\bf W}_{t\boldsymbol{:},j}^{(g)}) = \eta ||\text{\bf W}_{t\boldsymbol{:},j}^{(g)}||_2$, the proximal operator is given~by
\begin{equation*}
    \text{prox}_{\lambda^s g} (\text {\bf W}_{t\boldsymbol{:},j}^{(g)}) = \frac{\text {\bf W}_{t\boldsymbol{:},j}^{(g)}}{||\text {\bf W}_{t\boldsymbol{:},j}^{(g)}||_2} \left(||\text {\bf W}_{t\boldsymbol{:},j}^{(g)}||_2 -  \lambda^s \eta \right)_+.
\end{equation*}
 $(\psi)_+ \triangleq \max(0,\psi)$~\citep{parikh14}. This operator reduces the norm of $\mathbf{W}_{t\boldsymbol{:},j}^{(g)}$ by $\lambda^s \eta$, and shrink all $\mathbf{W}_{t\boldsymbol{:},j}^{(g)}$ to zero with $||\mathbf{W}_{t\boldsymbol{:},j}^{(g)}||_2 \leq \lambda^s \eta$. 


This operator is superior than other Bayesian shrinkage approaches \citep{shin17} which typically give small but non-zero valued estimates.  We  use Adam~\citep{Kingma15} for the remaining neural network parameters: $\theta$ and $\phi$. See Alg.\ref{alg:vi} for ITM-VAE pseudocode.

\begin{algorithm}[tb]
   \caption{Collapsed VI for ITM-VAE}
   \label{alg:vi}
\begin{algorithmic}
   \STATE {\bfseries Input:} data $\mathbf{x}^{(i)}$, sparsity parameter $\lambda$
   \STATE Let $\tilde{\mathcal{L}_t}$ be $\mathcal{L}(\phi_t, \theta_t, \mathcal{W}_t)$ but without $- \lambda \sum_{g,j} || \mathbf{W}_{t\boldsymbol{:},j}^{(g)} ||_2$.
   \REPEAT
   \STATE For each time point t
   \STATE Calculate $\nabla_{\phi_t} \tilde{\mathcal{L}_t}$, $\nabla_{\theta_t} \tilde{\mathcal{L}_t}$, and $\nabla_{\mathcal{W}_t} \tilde{\mathcal{L}_t}$.
   \STATE Update $\phi_t$ and $\theta_t$ with Adam optimizer.
   \STATE Let $\mathcal{W}_{t+1} = \mathcal{W}_t{t} - \eta \nabla_\mathcal{W} \tilde{\mathcal{L}_t}$.
   \FORALL{groups $g$, $j=1$ {\bfseries to} $K$}
       \STATE Set $\mathbf{W}_{t\boldsymbol{:},j}^{(g)} \gets \frac{\mathbf{W}_{t\boldsymbol{:},j}^{(g)}}{||\mathbf{W}_{t\boldsymbol{:},j}^{(g)}||_2} \left(||\mathbf{W}_{t\boldsymbol{:},j}^{(g)}||_2 - \eta \lambda \right)_+ $
   \ENDFOR
   \UNTIL{convergence in both $\sum_{t=1}^{T}\hat{\mathcal{L}_t}$ and $- \lambda \sum_{t=1}^{T}\sum_{g,j} || \mathbf{W}_{t\boldsymbol{:},j}^{(g)} ||_2$}
\end{algorithmic}
\end{algorithm}


\begin{figure*}[t!]
\begin{center}
\centerline{\includegraphics[width=5in]{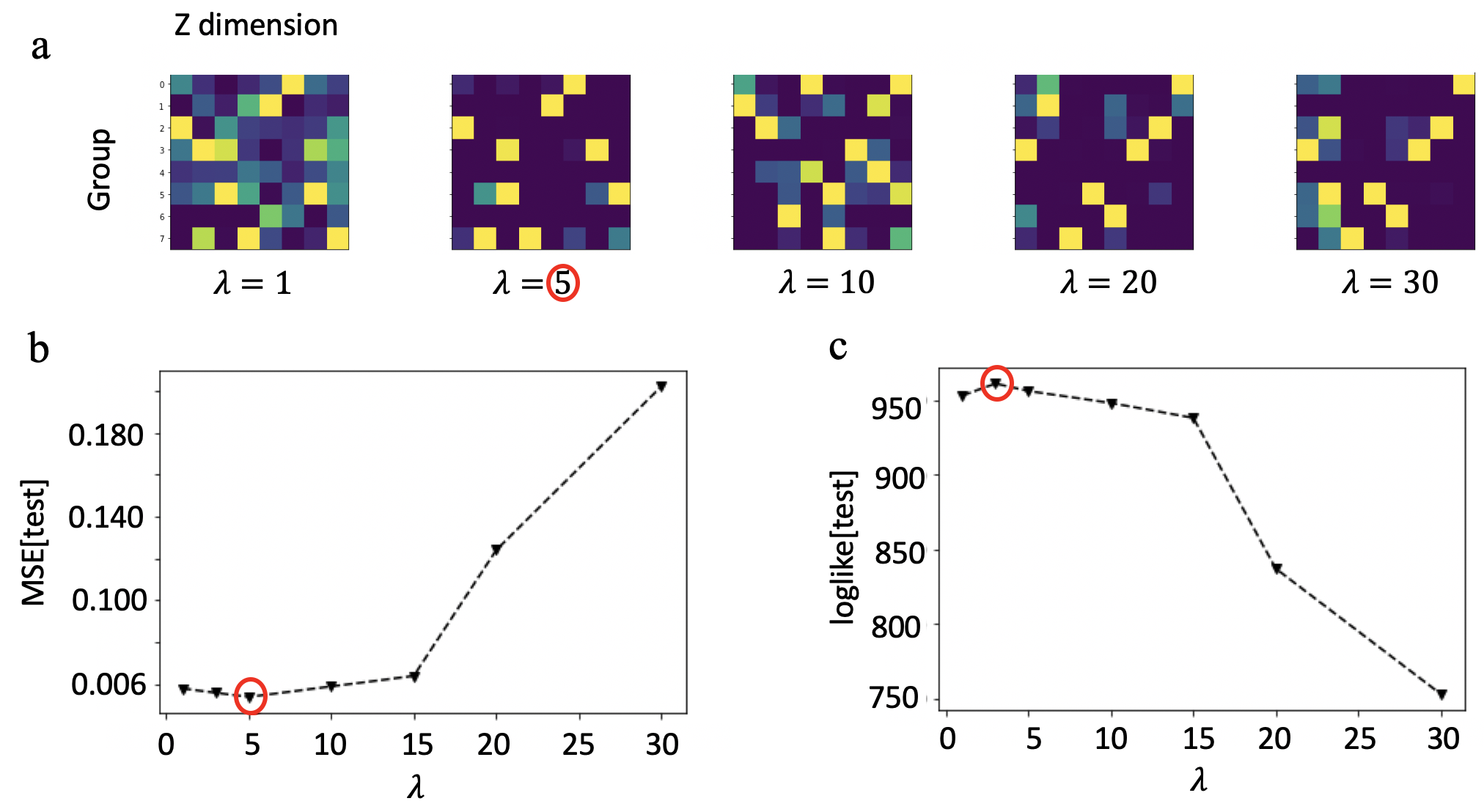}}
\caption{Artificial data experiments.   (a) The learned $\mathbf{W}_{t\boldsymbol{:},j}^{(g)}$ at time point $t=8$ for different $\lambda$ values, different rows represent different groups, yellow represents the dominant factor; (b) The mean squared error on the test set (MSE[test]) calculated on the concatenated time points on different $\lambda$ values (left); The loglikelihood value on the test set for different $\lambda$ values (right).}
\label{fig_mse}
\end{center}
\vskip -0.2in
\end{figure*}

\begin{figure*}[t!]
\begin{center}
\centerline{\includegraphics[width=5in]{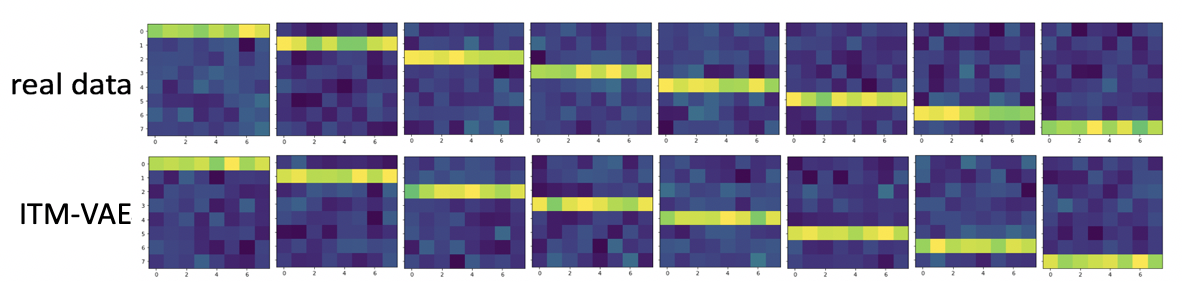}}
\caption{Real artificial data and reconstructed images from ITM-VAE.}
\label{fig:recon}
\end{center}
\vskip -0.2in
\end{figure*}

\section{Experiments}

\subsection{Methods Considered}
In addition to the ITM-VAE\footnote{Source code is available at https://github.com/lquvatexas/ITM-VAE}, we also consider VAE~\citep{Kingma14}, adapted conditional VAE (CVAE)~\citep{Sohn15}, oi-VAE~\citep{Ainsworth18}, VRNN~\citep{Chung15}, and GP-VAE~\citep{fortuin20}. VAE and oi-VAE: We concatenate data across different timesteps and treat the concatenated data independent. CVAE: CVAE requires the label information  as input to both the encoder and decoder networks. For our unsupervised problem, we assign different $t$ for the data as its label information. More specifically, the input for encoder is [$X_t$,t] and the input for decoder is [$Z_t$,t].

\begin{table*}[!t]
\caption{MSE from testing on the artificial data, motion capture data, and metabolomic data. The values are the means and respective standard errors.} \label{table:mse}
\centering
\begin{tabular}{lccr}
\hline 
Model &    Artificial Data & Motion Capture Data & Metabolomic Data           \\
\hline
VAE   & $0.029 \pm 0.002$    &        $0.090 \pm 0.016$   &  $0.072 \pm 0.048$\\
CVAE   & $0.003 \pm 0.001$    &             $0.053 \pm 0.004$ & $0.071 \pm 0.046$ \\
oi-VAE  & $0.005 \pm 0.001$   &            $0.057 \pm 0.006$  & $0.063 \pm 0.027$\\
VRNN    & $0.003 \pm 0.001$     &          $0.052\pm 0.004$  & $0.052 \pm 0.032$\\
GP-VAE & $0.004 \pm 0.001$      &          $0.045 \pm 0.006$   &$0.047 \pm 0.042$\\
\hline
ITM-VAE (ours)   & $\mathbf{0.002 \pm 0.001}$      &           $\mathbf{0.033 \pm 0.006}$ & $\mathbf{0.026 \pm 0.007}$\\
\hline
\end{tabular}
\end{table*}
\textbf{Evaluation metrics} To check and validate how well the disentanglement is achieved, we propose to visualize the $\mathbf{W}^{(g)}_{t\boldsymbol{:}, j}$ matrix at different timesteps $t$ and quantitatively compare the MSE[test] (mean squared error on the test data) with alternative methods. 

\textbf{Selection on $\lambda$ and $k$}
The parameter $\lambda$ controls the model sparsity, larger $\lambda$ will imply more column-wise sparsity in $\mathbf{W}_{t\boldsymbol{:},j}^{(g)}$, we propose to select $\lambda$ based on the learned $\mathbf{W}_{t\boldsymbol{:},j}^{(g)}$ ({Fig.~\ref{fig_mse}a) to check the sparsity and the MSE[test] ({Fig.~\ref{fig_mse}b). The latent dimension $k$ is chosen based on interpretation purpose.

\subsection{Artificial Data}

\textbf{Setup}  In order to visualize the performance, we generate one-bar images. The row position of the bar was taken as different time point labels, starting from the first row as time point $t=1$ to the last row as time point $t=8$. Thus, there are in  total $T=8$ time points. 

\textbf{Dataset Generation} We generate $2000$ $8\times8$  one-bar images and we add normal random noises with mean $0$ and standard deviation $0.05$ to the entire image. We randomly select 80\% (n=1600) of the image for training, 10\% 
(n=200) for validation and 10\% (n=200) for testing. For each batch, we use batch size $64$. Therefore, the data structure is of $8\times64\times64$ for ITM-VAE. In order to associate each dimension of $\textbf{z}$ with a unique row in the image, we chose $k=8$.

\textbf{Results} For ITM-VAE and CVAE, we calculate the MSE[test] on the concatenated time points of test data which is the same for the rest of the  experiments. The results are in {Fig.~\ref{fig:recon}} and ITM-VAE yields  the lowest MSE[test] comparing to other competitors {Table.~\ref{table:mse}}. 
VRNN and CVAE perform similarly and close to ITM-VAE. However, GP-VAE and oi-VAE do not perform very well. VAE has the largest MSE. These results indicate that ITM-VAE can learn multivariate time series data and reconstruct them very well. 
Meanwhile, we randomly select 64 images for each batch and replicate each image 20 times ($T=20$) to represent the perfect time series structure, the data structure is of $20\times64\times64$.  We showed the learned $\mathbf{W}_{t\boldsymbol{:},j}^{(g)}$ at time point $t=8$ for different $\lambda$ values in Fig.~\ref{fig_mse}a. It is clear that under $\lambda=5$ ITM-VAE  can successfully disentangle each of the dimensions of $\mathbf{z}$ to correspond to exactly one row (group) of the image at each time point.

\subsection{Motion Capture Data}
\textbf{Setup}  We consider the motion capture data obtained from CMU (http://mocap.cs.cmu.edu) to evaluate ITM-VAE's ability to handle complex longitudinal multivariate data. We use subject 7 data, which contains 11 trials of standard walking and one brisk walking recordings from the same person. For each trial, it contains different time frames of the person's moving skeleton, and it measures 59 joint angles split across 29 distinct joints. In this setting, we treat each distinct joint as a view, and each joint has 1 to 3 observed degrees of freedom to represent the different group dimensions. The task is to evaluate the model's ability for sensible dynamic disentanglement, model interpretability and generalization ability.

\textbf{Data} For model training, we use the data from 1 to 10 trials. In total, the 10 trials training data have 3776 frames. For testing, we use the $11$th trial data which has 315 frames. We set $T=32$ to train the model with batch size 32. For each batch, the data structure is of $32\times32\times59$ for ITM-VAE.

\textbf{Results} To check different latent dimensional effects of $\mathbf{z}$, we train ITM-VAE on $k=4,8,$ and $16$. {Fig.~\ref{fig:momo_recon}} shows the results for $k=8$.  ITM-VAE displays lower MSE[test] than all the other competitors in {Table.~\ref{table:mse}}. From the MSE[test], we can see that CVAE, VRNN, and oi-VAE perform very similarly with each other. This demonstrates that even oi-VAE is not a temporal model, its model structure can still capture the complex variations and our ITM-VAE is a temporal extension of oi-VAE.

To evaluate the generative ability of ITM-VAE, we show the reconstructed images of trial 11 in {Fig.~\ref{fig:momo_recon}} bottom row. The hidden dynamic information extracted from ITM-VAE generates very natural poses of human walking. In fact, there is clearly a moving pattern from the head to foot between neighboring timesteps. On the other hand, 
the results obtained from oi-VAE, which treats each time frame data independently, are very similar among each other, and there is no obvious trend in CVAE either. Since GP-VAE cannot capture the shared temporal structure, it fails to capture some details of the trend in foot and hand. We further compare the test-loglikehood between ITM-VAE and oi-VAE on trial 11 and trial 12, which is the brisk walk data. Table.~\ref{table1} records the log-likelihood for both ITM-VAE and oi-VAE models on two testing trials with $k=4,8$. ITM-VAE has higher test log-likelihood and both methods achieve higher test log-likelihood when the latent dimension $k$ is larger. This indicates that ITM-VAE can achieve better generalization than oi-VAE, because the brisk walking trial is very different from the training walking trials. 

\begin{figure}[t!]
\vskip 0.1in
\begin{center}
\centerline{\includegraphics[width=4in]{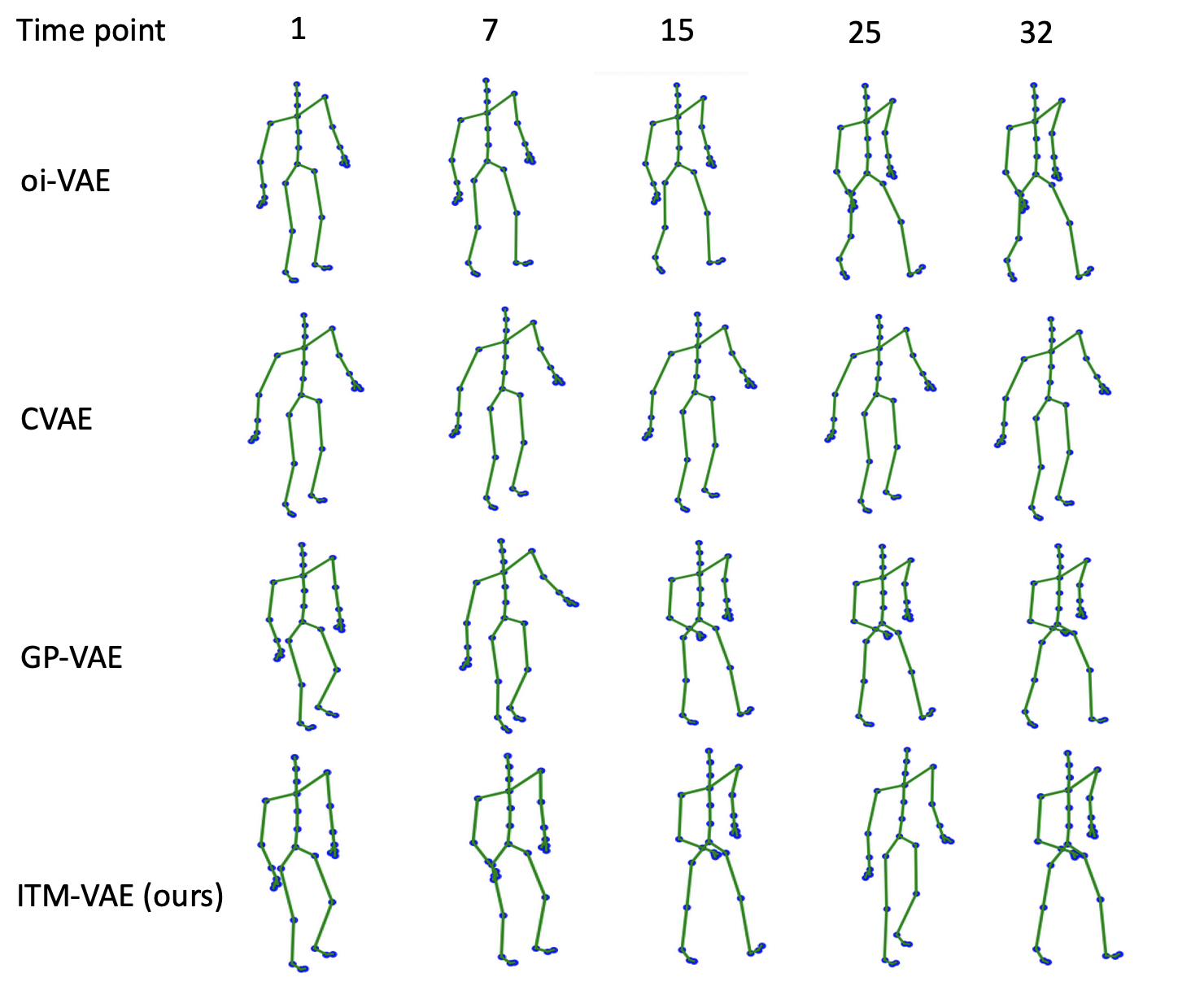}}
\caption{Reconstructed images from motion capture data. All the models are trained on the first 10 trial walking data. The generated images are from trial 11 (different from training) at time point $t=1,7,15,25,32$.} 
\label{fig:momo_recon}
\end{center}
\vskip -0.4in
\end{figure}

Fig.~\ref{fig:momo} shows that the factors change across different time points. For example, from time point 1 to 3, the first factor (first column of the left and middle images) changes from lfoot (left foot) to rfoot (right foot), factor 2 changes from rwrist (right wrist) to thorax, and factor 7 changes from rwrist to rtibia (right tibia). These changes are indeed reasonable because when we start to walk with the foot, the tibia and the thorax move accordingly~\citep{Versichele12}. The above observation demonstrates that the learned latent representation from ITM-VAE has an intuitive anatomical interpretation for different time points. We also provided a detailed list of the joints per latent variable dimension that are most strongly influenced by each factor in {Table.~\ref{table2}}. For example, factor 1 represents foot and lower back, factor 2 represents wrist, thorax and upper back, and factor 8 represents wrist, foot and hand.
All these observations demonstrate that ITM-VAE can track the dynamic latent embeddings and provide meaningful  interpretation.

\begin{figure}[t!]
\vskip 0.1in
\begin{center}
\centerline{\includegraphics[width=4in]{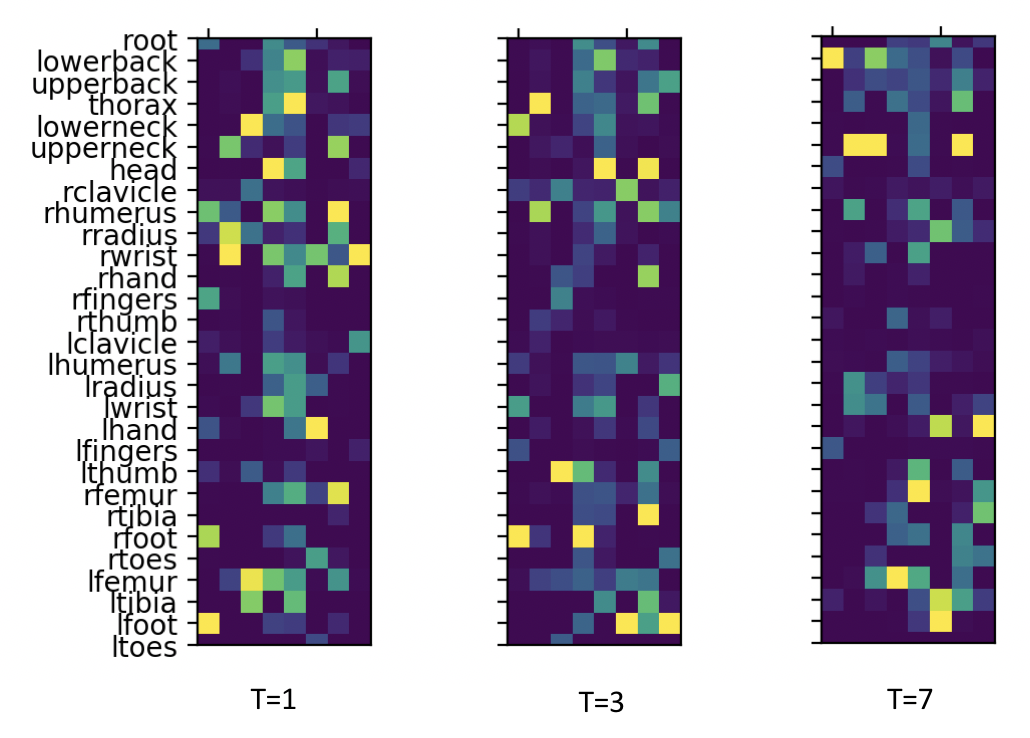}}
\caption{Learned ITM-VAE $\mathbf{W}_{t\boldsymbol{:},j}^{(g)}$ at time points $t=1,3,7$ under $k=8$. Each row corresponds to each group of the joints, columns represent different latent dimensions. Specifically, the values of latent dimensions are color-coded from dark blue (zero) to yellow (maximum non-zero value) to indicate the strength of the latent-to-group mappings $\mathbf{W}_{t\boldsymbol{:},j}^{(g)}$.} 
\label{fig:momo}
\end{center}
\vskip -0.2in
\end{figure}

\begin{table}[t!]
\tiny
\caption{ITM-VAE results on motion capture data interpretation. Top 3 joints corresponding to each latent dimension determined by $\mathbf{W}^{(g)}_{\boldsymbol{\cdot}, j}$ in Fig. \ref{fig:momo}. k represents the latent dimension, T indicates the time points.}
\label{table2}
\begin{center}
\begin{tabular}{lcccr}
\toprule
 k & T=1 & T=3 & T=7 \\
\midrule
1    & left foot, right foot, right fingers & right foot, lower neck, left wrist & lower back, left fingers, head\\
2    & right wrist, right radius, upper neck& thorax, right humerus, right foot & upper back, left radius, left wrist\\
3   & lower neck, left femur, left tibia & left thumb, right clavicle, right fingers & upper neck, lower back, left femur \\
4    & head, right humerus, left femur& right foot, left thumb, left wrist & left femur, right tibia, thorax\\
5     & thorax, lower back, left tibia & head, lower back, left tibia & right femur, left thumb, left femur\\
6     & left hand, right wrist, right toes & left foot, right clavicle, left femur &  left foot, left tibia, left hand\\
7     & right humerus, right femur, right hand  & head, right tibia, right hand &  upper neck, thorax, left tibia\\
 
8     & right wrist, left clavicle, lower neck  & left foot, left radius,  upper back&  left hand, right tibia, right femur\\
\bottomrule
\end{tabular}
\end{center}
\vskip -0.3in
\end{table}

\subsection{Metabolomic Data}
\textbf{Setup} In this section, we propose to analyze the data obtained from a longitudinal 
study~\citep{Jozefczuk10}, where one of the objectives is to compare metabolic changes of \textit{E.coli} response to five different perturbations: cold, heat, oxidative stress, lactose diauxie, and stationary phase. The task is to evaluate the model on limited sample size studies which is common in the life sciences field.

\begin{table}[t]
\caption{Test log-likelihood for ITM-VAE and oi-VAE trained on the first 10 trials of walking data. Table includes results for a  test walk (similar as training) and the brisk walk trial (different from training).}
\label{table1}
\vskip 0.15in
\begin{center}
\begin{sc}
\begin{tabular}{lccr}
\toprule
 & Standard Walk & Brisk Walk \\
\midrule
ITM-VAE (k=4)    & $\mathbf{-93,221}$ & $\mathbf{-30,056}$& \\
oi-VAE (k=4) & -1,006,120 & -598,660& \\
ITM-VAE (k=8)    &$\mathbf{-17,667}$ & $\mathbf{-36,299}$&  \\
oi-VAE (k=8)    & -998,849 &  -492,411&         \\
\bottomrule
\end{tabular}
\end{sc}
\end{center}
\end{table}

\begin{figure}[t!]
\begin{center}
\centerline{\includegraphics[width=4in]{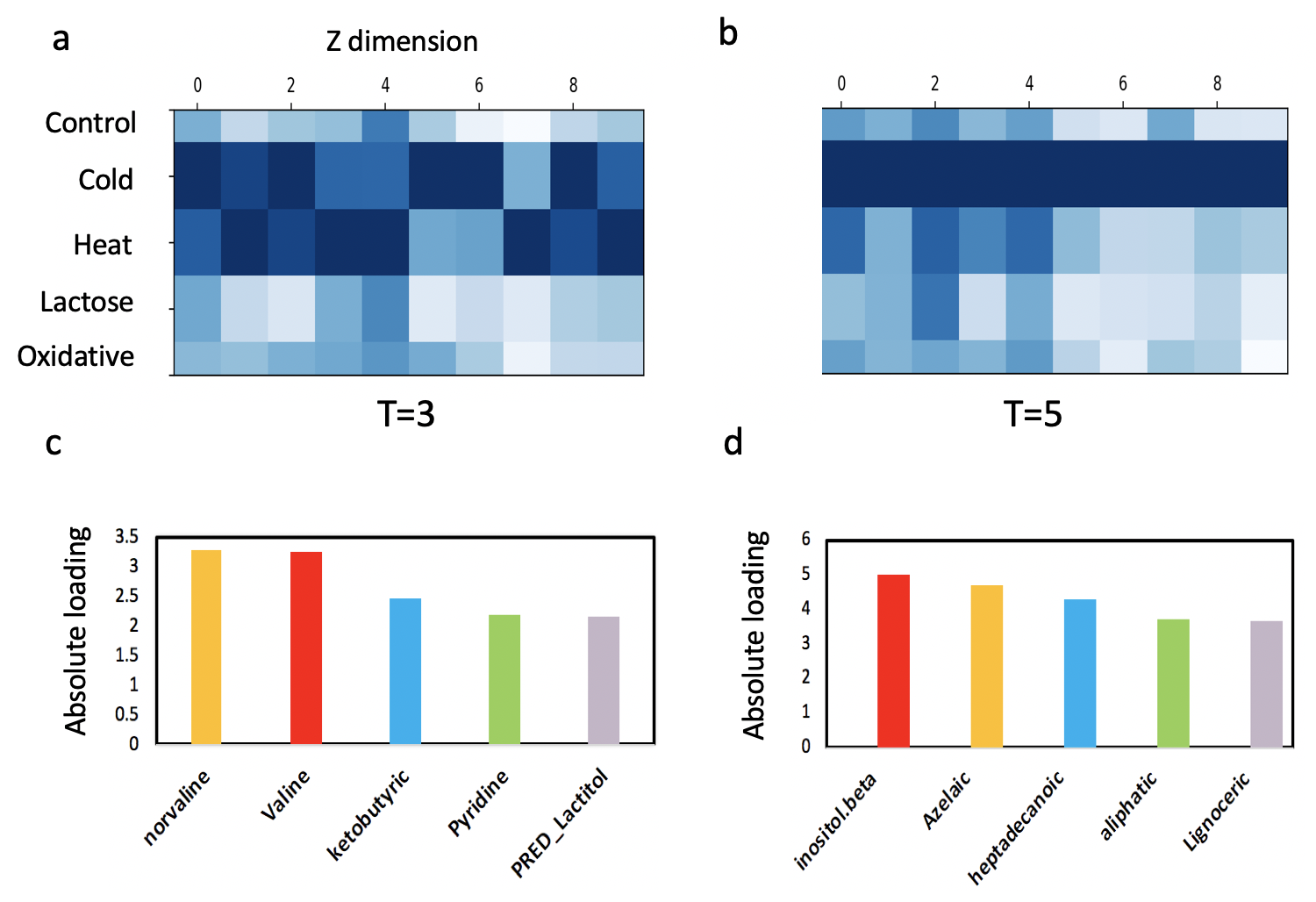}}
\caption{Results on metabolomic data with $k=10$. {(a)}-{(b)} Learned $\mathbf{W}_{t\boldsymbol{:},j}^{(g)}$ at time point $t=3,5$ under cold, heat, lactose shift and oxidative stress from ITM-VAE (blue represents dominant factor); {(c)} Absolute loading on factor 5 at $t=3$ under heat group; {(d)} Absolute loading on factor 10 at $t=5$ under cold group. }
\label{fig:metabo}
\end{center}
\vskip -0.3in
\end{figure}

\textbf{Data} The dataset contains 196 metabolite expression values measured for 8 subjects at 12 different time points under five stress conditions. We treat each condition as a group and randomly select 6 subjects as the training set and the remaining 2 subjects as the test set. We use batch size $n=2$, and for each batch, the data structure is of $12\times2\times980$ for ITM-VAE. 

\textbf{Results} ITM-VAE has the lowest MSE[test] in {Table.~\ref{table:mse}} (last column). GP-VAE and VRNN perform better than CVAE and other non-temporal competitors. This result further demonstrates that the generative model structure of ITM-VAE and oi-VAE can handle the limited sample size problems better than traditional VAEs, because oi-VAE is non-temporal model so it performs worse than VRNN and GP-VAE. The learned group-weights $\mathbf{W}_{t\boldsymbol{:},j}^{(g)}$ from ITM-VAE are shown in {Fig.~\ref{fig:metabo}a-b}. For ITM-VAE, it is clear that at time $t=3$, most of the factors' variations are explained by cold and heat groups, at time $t=5$, cold group explains most of the variations.  
These results are consistent with the findings in the original paper~\citep{Jozefczuk10}. 
Another important downstream analysis is the inspection of top features with largest weight: The loadings can give insights into the biological process underlying the heterogeneity captured by a latent factor. There might be scale differences among groups, the weights of different views are not directly comparable. For simplicity, we scale each weight vector by its absolute value. In Fig. \ref{fig:metabo}c and d, we plotted the top 5 metabolites with largest absolute weight of two interesting factors. Like the top feature Norvaline in Fig. \ref{fig:metabo}c is known to promote tissue regeneration and muscle growth \citep{ming09}, and to become a precursor in the penicillin biosynthetic pathway. We leave more detailed interpretation for biological interest. 


\section{Discussion}
We develop an interpretable nonlinear framework for temporal multi-view data, namely ITM-VAE, with the goal of disentangling the dynamically shared latent embeddings for the complex multi-view varations. One key feature of ITM-VAE is its ability to integrate the VRNN to the shared latent variables among different groups in order to model the complex sequence data and extract the dependency relationships. 


Our empirical analyses on both motion capture and metabolomics data demonstrate that ITM-VAE can successfully  extract the hidden time dependence structures. 
More importantly, the achieved model efficiency and interpretability does not occur at the cost of model generalization. Because ITM-VAE can model complex temporal multi-view data and result in interpretable results, we believe ITM-VAE will have wide applications in different fields.

\bibliography{paper}

\begin{thebibliography}{30}
\providecommand{\natexlab}[1]{#1}
\providecommand{\url}[1]{\texttt{#1}}
\expandafter\ifx\csname urlstyle\endcsname\relax
  \providecommand{\doi}[1]{doi: #1}\else
  \providecommand{\doi}{doi: \begingroup \urlstyle{rm}\Url}\fi

\bibitem[Ainsworth et~al.(2018)Ainsworth, Foti, Lee, and Fox]{Ainsworth18}
S.~K. Ainsworth, N.~J. Foti, A.~K.~C. Lee, and E.~B. Fox.
\newblock oi-vae: Output interpretable vaes for nonlinear group factor
  analysis.
\newblock In \emph{Proceedings of the 35th International Conference on Machine
  Learning (ICML 18)}, 2018.

\bibitem[Archer et~al.(2015)Archer, Park, L., Cunningham, and
  Paninski]{Archer15}
E.~Archer, I.~M. Park, L., J.~Cunningham, and L.~Paninski.
\newblock Black box variational inference for state space models.
\newblock \emph{arXiv preprint arXiv:1511.07367}, 2015.

\bibitem[Casale et~al.(2018)Casale, Dalca, Saglietti, Listgarten, and
  Fusi]{casale18}
F.~P. Casale, A.~V. Dalca, L.~Saglietti, J.~Listgarten, and N.~Fusi.
\newblock Gaussian process prior variational autoencoders.
\newblock In \emph{Proceedings of the 35th International Conference on Machine
  Learning (ICML 18)}, 2018.

\bibitem[Cho et~al.(2014)Cho, van Merrienboer, Gulcehre, Bahanau, Bougares,
  Schwenk, and Bengio]{cho14}
K.~Cho, B.~van Merrienboer, C.~Gulcehre, D.~Bahanau, F.~Bougares, H.~Schwenk,
  and Y.~Bengio.
\newblock Learning phrase representations using {RNN} encoder{--}decoder for
  statistical machine translation.
\newblock In \emph{Proceedings of the 2014 Conference on Empirical Methods in
  Natural Language Processing ({EMNLP})}, pages 1724--1734, Doha, Qatar,
  October 2014. Association for Computational Linguistics.
\newblock \doi{10.3115/v1/D14-1179}.

\bibitem[Chung et~al.(2015)Chung, Kastner, Dinh, Goel, and Courville]{Chung15}
J.~Chung, K.~Kastner, L.~Dinh, K.~Goel, and A.~Courville.
\newblock A recurrent latent variable model for sequential data.
\newblock In \emph{Proceedings of the 28th International Conference on Neural
  Information Processing Systems (NIPS 15)}, pages 2980--2988, 2015.

\bibitem[Fortuin et~al.(2020)Fortuin, Baranchuk, Raetsch, and Mandt]{fortuin20}
V.~Fortuin, D.~Baranchuk, G.~Raetsch, and S.~Mandt.
\newblock Gp-vae: Deep probabilistic time series imputation.
\newblock In \emph{Proceedings of the Twenty Third International Conference on
  Artificial Intelligence and Statistics (AISTATS 20)}, volume 108, pages
  1651--1661, 2020.

\bibitem[Gao et~al.(2013)Gao, Brown, and Engelhardt]{Gao13}
C.~Gao, C.~D. Brown, and B.~E. Engelhardt.
\newblock A latent factor model with a mixture of sparse and dense factors to
  model gene expression data with confounding effects.
\newblock \emph{arXiv preprint arXiv:1310.4792}, 2013.

\bibitem[Graves(2013)]{Graves13}
A.~Graves.
\newblock Generating sequences with recurrent neural networks.
\newblock \emph{arXiv preprint arXiv:1308.0850}, 2013.

\bibitem[Hermans and Schrauwen(2013)]{Hermans13}
M.~Hermans and B.~Schrauwen.
\newblock Training and analyzing deep recurrent neural networks.
\newblock In \emph{Proceedings of the 26th International Conference on Neural
  Information Processing Systems (NIPS 13)}, volume~1, pages 190--198, 2013.

\bibitem[Hochreiter and Schmidhuber(1997)]{hochreiter97}
S.~Hochreiter and J.~Schmidhuber.
\newblock Long short-term memory.
\newblock \emph{Neural computation}, 9\penalty0 (8):\penalty0 1735--1780, 1997.

\bibitem[Jozefczuk et~al.(2010)Jozefczuk, Klie, Catchpole, Szymanski,
  Cuadros-Inostroza, Steinhauser, Selbig, and Willmitzer]{Jozefczuk10}
S.~Jozefczuk, S.~Klie, G.~Catchpole, J.~Szymanski, A.~Cuadros-Inostroza,
  D.~Steinhauser, J.~Selbig, and L.~Willmitzer.
\newblock Metabolomic and transcriptomic stress response of escherichia coli.
\newblock \emph{Molecular Systems Biology}, 6, 2010.

\bibitem[Kingma and Ba(2015)]{Kingma15}
D.~P. Kingma and J.~L. Ba.
\newblock Adam: A method for stochastic optimization.
\newblock \emph{arXiv preprint arXiv:1412.6980v9}, 2015.

\bibitem[Kingma and Welling(2014)]{Kingma14}
D.~P. Kingma and M.~Welling.
\newblock Auto-encoding variational bayes.
\newblock \emph{arXiv preprint arXiv:1312.6114}, 2014.

\bibitem[Klami et~al.(2015)Klami, Virtanen, Leppaaho, and Kaski]{Klami15}
A.~Klami, S.~Virtanen, E.~Leppaaho, and S.~Kaski.
\newblock Group factor analysis.
\newblock \emph{IEEE Transactions on Neural Networks and Learning Systems},
  26:\penalty0 2136--2147, 2015.

\bibitem[Kyung et~al.(2010)Kyung, Gill, and Casella]{Kyung10}
M.~Kyung, J.~Gill, and G.~Casella.
\newblock Penalized regression, standard errors, and bayesian lassos.
\newblock \emph{Bayesian Analysis}, 5\penalty0 (2):\penalty0 369--412, 2010.

\bibitem[Lawrence et~al.(2019)Lawrence, Ek, and Campbell]{Lawrence19}
A.~R. Lawrence, C.~H. Ek, and N.~D.F. Campbell.
\newblock Dp-gp-lvm: A bayesian non-parametric model for learning multivariate
  dependency structures.
\newblock In \emph{Proceedings of the 36th International Conference on Machine
  Learning (ICML 19)}, 2019.

\bibitem[Leppaaho et~al.(2017)Leppaaho, Ammad{--}ud{--}din, and
  Kaski]{Leppaaho17}
E.~Leppaaho, M.~Ammad{--}ud{--}din, and S.~Kaski.
\newblock Gfa: Exploratory analysis of multiple data sources with group factor
  analysis.
\newblock \emph{Journal of Machine Learning Research}, 18:\penalty0 1--5, 2017.

\bibitem[Lucas et~al.(2010)Lucas, Kung, and Chi]{Lucas10}
J.~E. Lucas, H.~Kung, and J.~A. Chi.
\newblock Latent factor analysis to discover pathway-associated putative
  segmental aneuploidies in human cancers.
\newblock \emph{PLoS Computational Biology}, 6\penalty0 (9):\penalty0 e1000920,
  2010.

\bibitem[Martens and Sutskever(2011)]{Martens11}
J.~Martens and I.~Sutskever.
\newblock Learning recurrent neural networks with hessian-free optimization.
\newblock In \emph{Proceedings of the 28th International Conference on
  International Conference on Machine Learning (ICML 11)}, pages 1033--1040,
  2011.

\bibitem[Ming et~al.(2009)Ming, Rajapakse, Carvas, Ruffieux, and Yang]{ming09}
X.~F. Ming, A.~G. Rajapakse, J.~M. Carvas, J.~Ruffieux, and Z.~H. Yang.
\newblock Inhibition of s6ki accounts partially for the anti-inflammatory
  effects of the arginase inhibitor l-norvaline.
\newblock \emph{BMC Cardiovascular Disorders}, 9\penalty0 (12), 2009.

\bibitem[Parikh and Boyd(2014)]{parikh14}
N.~Parikh and S.~Boyd.
\newblock Proximal algorithms.
\newblock \emph{Foundations and Trends{\textregistered} in Optimization},
  1\penalty0 (3):\penalty0 127--239, 2014.

\bibitem[Pascanu et~al.(2013)Pascanu, Mikolov, and Bengio]{Pascanu13}
R.~Pascanu, T.~Mikolov, and Y.~Bengio.
\newblock On the difficulty of training recurrent neural networks.
\newblock In \emph{Proceedings of the 30th International Conference on
  International Conference on Machine Learning (ICML 13)}, volume~28, pages
  1310--1318, 2013.

\bibitem[Pournara and Wernisch(2007)]{Pournara07}
I.~Pournara and L.~Wernisch.
\newblock Factor analysis for gene regulatory networks and transcription factor
  activities profiles.
\newblock \emph{BMC Bioinformatics}, 8:\penalty0 61, 2007.

\bibitem[Rezende et~al.(2014)Rezende, Mohamed, and Wierstra]{rezende14}
D.~J. Rezende, S.~Mohamed, and D.~Wierstra.
\newblock Stochastic backpropagation and approximate inference in deep
  generative models.
\newblock In \emph{Proceedings of the 31th International Conference on Machine
  Learning (ICML 14)}, 2014.

\bibitem[Shin(2017)]{shin17}
M.~Shin.
\newblock \emph{Priors for bayesian shrinkage and high-dimensional model
  selection}.
\newblock PhD thesis, College Station TX, 2017.

\bibitem[Sohn et~al.(2015)Sohn, Lee, and Yan]{Sohn15}
K.~Sohn, H.~Lee, and X.C. Yan.
\newblock Learning structured output representation using deep conditional
  generative models.
\newblock In \emph{Proceedings of the 26th International Conference on Neural
  Information Processing Systems (NIPS 15)}, pages 3483--3491, 2015.

\bibitem[Sridharan and Kakade(2008)]{Sridharan08}
K.~Sridharan and S.~M. Kakade.
\newblock An information theoretic framework for multi-view learning.
\newblock In \emph{In Proceedings of COLT}, pages 403--414, 2008.

\bibitem[Versichele et~al.(2012)Versichele, Neutens, Delafontaine, and
  Weghe]{Versichele12}
M.~Versichele, T.~Neutens, M.~Delafontaine, and N.~V.~D. Weghe.
\newblock The use of bluetooth for analysing spatiotemporal dynamics of human
  movement at mass events: A case study of the ghent festivities.
\newblock \emph{Applied Geography}, 32:\penalty0 208--220, 2012.

\bibitem[Yang et~al.(2017)Yang, Ramesh, Chitta, Madhvanath, Bernal, and
  Luo]{yang17}
X.~Yang, P.~Ramesh, R.~Chitta, S.~Madhvanath, E.~A. Bernal, and J.~Luo.
\newblock Deep multimodal representation learning from temporal data.
\newblock In \emph{IEEE Conference on Computer Vision and Pattern Recognition
  (CVPR 17)}, 2017.

\bibitem[Zhao et~al.(2016)Zhao, Gao, Mukherjee, and Engelhardt]{zhao16}
S.~Zhao, C.~Gao, S.~Mukherjee, and B.~E. Engelhardt.
\newblock Bayesian group factor analysis with structured sparsity.
\newblock \emph{Journal of Machine Learning Research}, 17\penalty0
  (4):\penalty0 1--47, 2016.

\end{thebibliography}
\clearpage
\appendix
\section*{Appendix }
\subsection*{A.1 Network Architecture}





\textbf{Feature extraction network}:
Since we introduced the random hidden state $\mathbf{h}_t$ for the recurrent neural network, we use neural networks $\varphi_{\tau}^{\mathbf{x}}$ and $\varphi_{\tau}^{\mathbf{z}}$ for feature extraction from $\mathbf{x}_t$ and $\mathbf{z}_t$, respectively.

\begin{itemize}
    \item $\varphi_{\tau}^{\mathbf{x}}(\mathbf{x}_t)= \mathbf{W}_1 \mathbf{x}_t + b_1$
    \item $\varphi_{\tau}^{\mathbf{z}}(\mathbf{z}_t) = \mathbf{W}_3 \text{relu} (\mathbf{W}_2 \mathbf{z} + b_2) + b_3$
\end{itemize}

After feature extraction from $\mathbf{x}_t$ and $\mathbf{z}_t$, then, we  stack $\mathbf{x}_t$ and $\mathbf{z}_t$ with $\mathbf{h}_{t-1}$ together for the inference and generative model respectively.\\

\textbf{Artificial data model structure}: 
\begin{itemize}
    \item Encoder:\\
    - $\mu(\mathbf{x}_t + \mathbf{h}_{t-1}) = \mathbf{W}_1 (\mathbf{x}_t + \mathbf{h}_{t-1}) + b_1$.\\
    - $\sigma(\mathbf{x}_t + \mathbf{h}_{t-1}) = \text{exp}(\mathbf{W}_2 (\mathbf{x}_t + \mathbf{h}_{t-1}) + b_2)$.\\
    \item Deconder:\\
    - $\mu(\mathbf{z}_t + \mathbf{h}_{t-1}) = \mathbf{W}_{3t} (\mathbf{z}_t + \mathbf{h}_{t-1}) + b_3$.\\
    - $\sigma(\mathbf{z}_t + \mathbf{h}_{t-1}) = \text{exp}(b_4)$.\\
\end{itemize}
\textbf{Motion capture data model structure}: 

\begin{itemize}
    \item Encoder:\\
    - $\mu(\mathbf{x}_t + \mathbf{h}_{t-1}) = \mathbf{W}_2\text{relu}(\mathbf{W}_1 (\mathbf{x}_t + \mathbf{h}_{t-1})) + b_1$.\\
    - $\sigma(\mathbf{x}_t + \mathbf{h}_{t-1})$ = \text{exp}($\mathbf{W}_3 \text{relu}(\mathbf{W}_1 (\mathbf{x}_t + \mathbf{h}_{t-1})) + b_2$).\\
    \item Deconder:\\
    - $\mu(\mathbf{z}_t + \mathbf{h}_{t-1}) = \mathbf{W}_{3t}  \text{tanh}(\mathbf{z}_t + \mathbf{h}_{t-1}) + b_3$.\\
    - $\sigma(\mathbf{z}_t + \mathbf{h}_{t-1})$ = exp($b_4$).\\
\end{itemize}


\textbf{Metabolomic data model structure}:
\begin{itemize}
    \item Encoder:\\
    - $\mu(\mathbf{x}_t + \mathbf{h}_{t-1}) = \mathbf{W}_2\text{relu}(\mathbf{W}_1 (\mathbf{x}_t + \mathbf{h}_{t-1})) + b_1$.\\
    - $\sigma(\mathbf{x}_t + \mathbf{h}_{t-1})$ = \text{exp}($\mathbf{W}_3 \text{relu}(\mathbf{W}_1 (\mathbf{x}_t + \mathbf{h}_{t-1})) + b_2$).\\
    \item Deconder:\\
    - $\mu(\mathbf{z}_t + \mathbf{h}_{t-1}) = \mathbf{W}_{3t}  \text{tanh}(\mathbf{z}_t + \mathbf{h}_{t-1}) + b_3$.\\
    - $\sigma(\mathbf{z}_t + \mathbf{h}_{t-1})$ = exp($b_4$).\\
\end{itemize}

\subsection*{A.2 Experimental Details}
 We ran Adam for the inference and generative net parameters optimization with learning rate 1$e$-3. Proximal gradient descent was run on $\mathbf{W}_t$ with learning rate 1$e$-4.  \textbf{Artificial data}: We  chose $\lambda = 5$. For the first part of experiment, we want to select $\lambda$ so we randomly selected 64 images at each iteration and replicate each image 20 times as one batch, we ran for 10,000 iterations. The data structure is $20\times64\times64$.
 For the second part of experiment, we assign the row position of bar as the time label, so in total we have $t=8$ different types of images, the data structure for each batch is $8\times64\times64$. \textbf{Motion capture data}: We chose $\lambda$ = 5, and we used $T=32$ frames and replicate each frame 32 times to stack as one batch ( $32\times32\times59$) to train our model, optimization was run for 100 epochs. \textbf{Metabolomic data}: We chose $\lambda=10$, we randomly selected $n=2$ as one batch, the data structure for each batch is $12\times2\times980$, we ran 10,000 epochs.
 

\subsection*{A.3 Supplementary Figures}

\begin{figure}[t!]
\vskip -0.05in
\begin{center}
\centerline{\includegraphics[width=1 \textwidth]{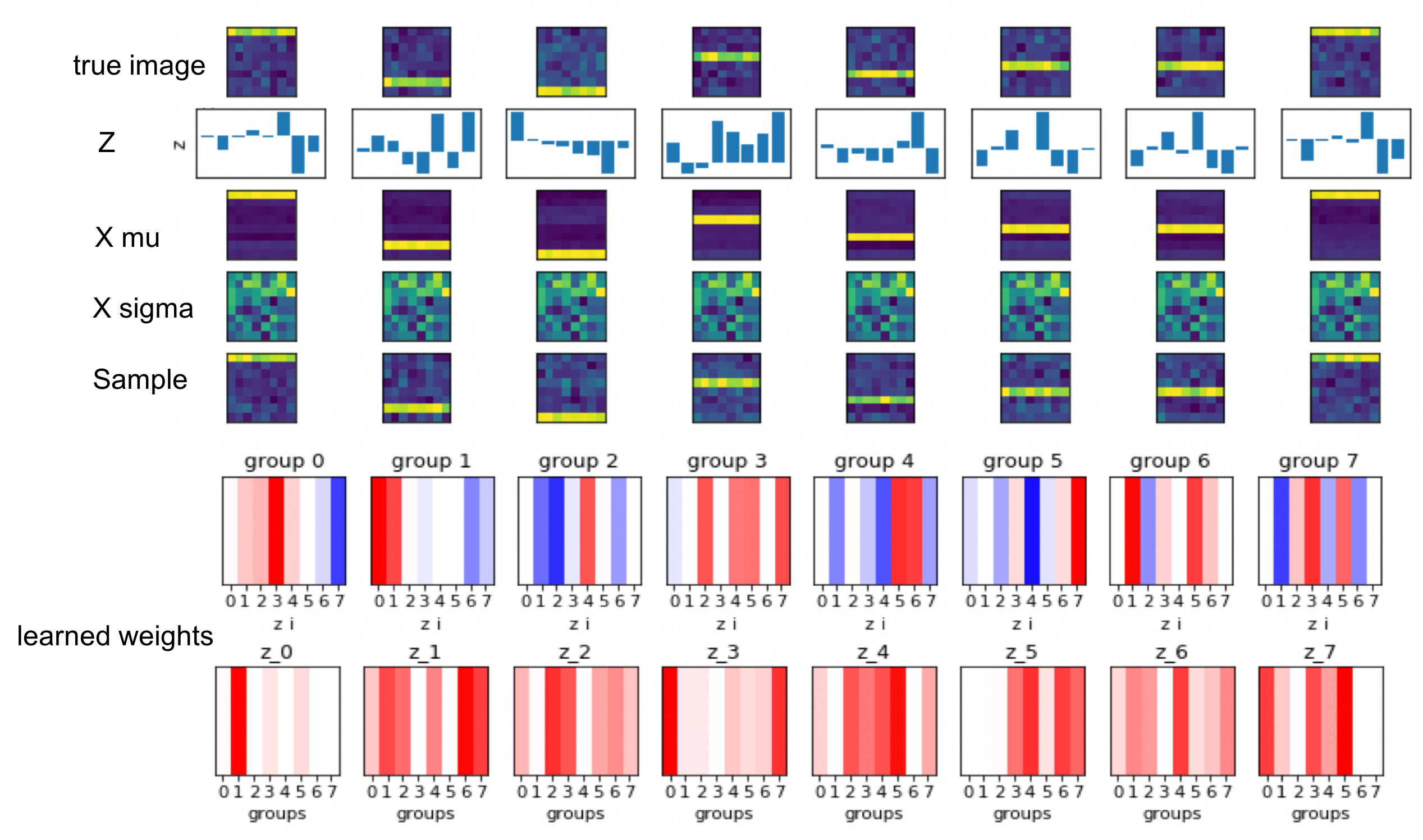}}
\caption{\textbf{Additional results of ITM-VAE on artificial data}. $k=8$, $\lambda=5$, iteration = 10000, batch-size = 64, $t=8$. Here, we didn't assign the $t$ label to each image since we want to check the model sparsity induced by $\lambda$. \textbf{true image}: the training image, \textbf{Z}: the sampled z values from encoder, \textbf{X mu}: the decoder mean, \textbf{X sigma}: the decoder sigma, \textbf{sample}: the reconstructed image from decoder mean and sigma, \textbf{learned weights}: the learned weights from the model.}

\label{fig:appendix_artificial}
\end{center}
\vskip -0.2in
\end{figure}

\begin{figure}[t!]
\vskip -0.05in
\begin{center}
\centerline{\includegraphics[width=1 \textwidth]{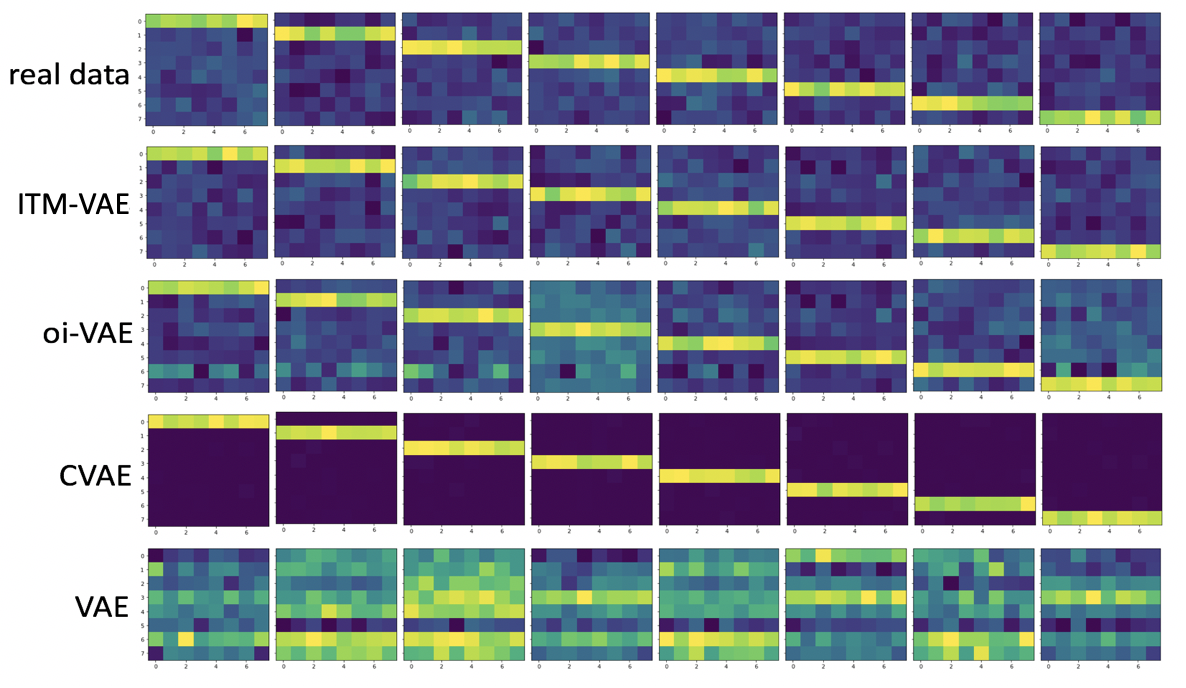}}
\caption{\textbf{Additional results of ITM-VAE on artificial data}. Reconstructed images.}

\label{fig:appendix_reconstructed}
\end{center}
\vskip -0.2in
\end{figure}

\begin{figure}[t!]
\vskip -0.05in
\begin{center}
\centerline{\includegraphics[width=1\textwidth]{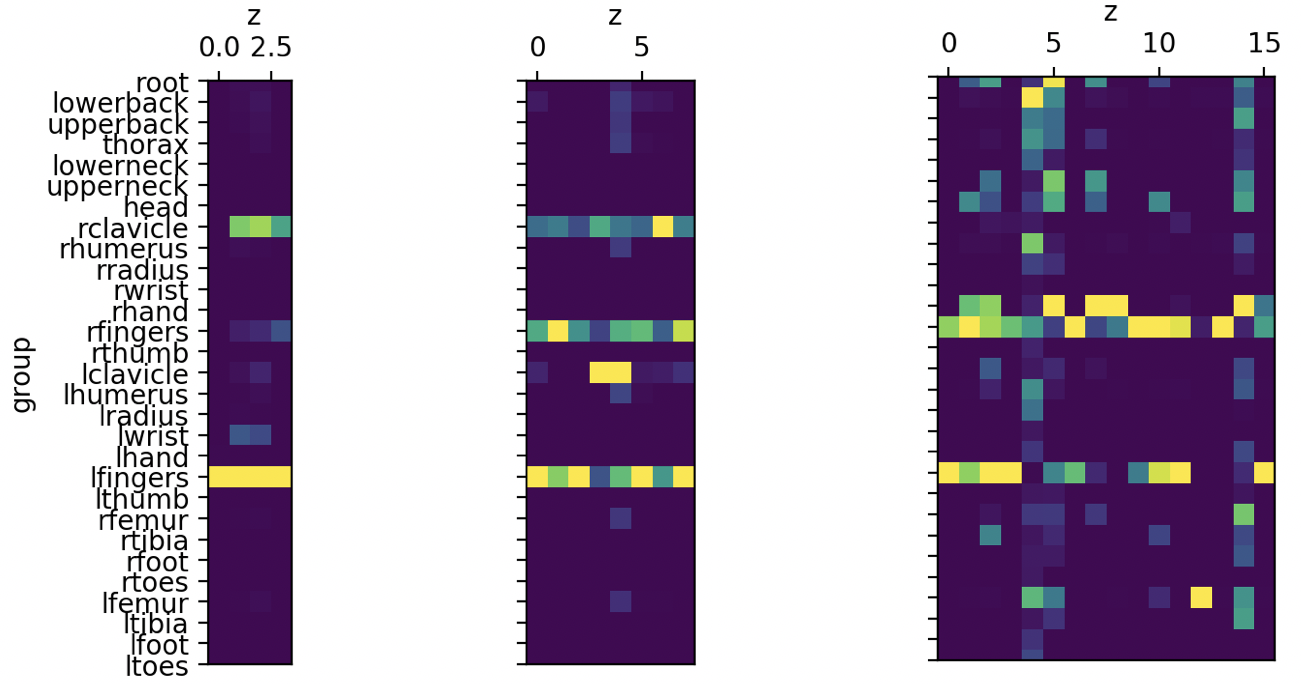}}
\caption{\textbf{Additional results from motion capture data.}  The learned $\mathbf{W}_{t\boldsymbol{:},j}^{(g)}$ at time point $t=10$ for $k=4$ (left), $k=8$ (middle), and $k=16$ (right) with $\lambda$ = 5. }
\label{fig:appendix_momo}
\end{center}
\vskip -0.2in
\end{figure}

\end{document}